\documentclass[letterpaper]{article} 
\usepackage[]{aaai23}  
\usepackage{times}  
\usepackage{helvet}  
\usepackage{courier}  
\usepackage[hyphens]{url}  
\usepackage{graphicx} 
\usepackage{natbib}  
\usepackage{caption} 
\usepackage{comment}
\usepackage[noend]{algpseudocode}
\usepackage{algorithm,algorithmicx}
\usepackage{xspace}
\usepackage{nicefrac}

\newcommand*\Let[2]{\State #1 $\gets$ #2}
\algrenewcommand\algorithmicrequire{\textbf{inputs:}}
\algrenewcommand\algorithmicensure{\textbf{outputs:}}

\newcommand{\round}[1]{\ensuremath{\lfloor#1\rceil}}
\algrenewcommand\algorithmicindent{1.0em}
\algrenewcommand\textproc{}
\algrenewcomment[1]{\(\triangleright\) #1}
\algnewcommand{\LineComment}[1]{\State \(\triangleright\) #1}

\usepackage{newfloat}
\usepackage{listings}
\DeclareCaptionStyle{ruled}{labelfont=normalfont,labelsep=colon,strut=off} 
\floatstyle{ruled}
\newfloat{listing}{tb}{lst}{}
\floatname{listing}{Listing}
\usepackage{multirow}
\usepackage{adjustbox}
\usepackage{amsfonts}
\usepackage{amssymb}
\usepackage{tabto}
\usepackage{makecell}
\usepackage{xcolor}
\usepackage{pifont}
\newcommand{\cmark}{\ding{51}}%
\newcommand{\xmark}{\ding{55}}%

\newcommand{\cc}[1]{{\color{blue} #1}}

\makeatletter
\def\thickhline{%
  \noalign{\ifnum0=`}\fi\hrule \@height \thickarrayrulewidth \futurelet
   \reserved@a\@xthickhline}
\def\@xthickhline{\ifx\reserved@a\thickhline
               \vskip\doublerulesep
               \vskip-\thickarrayrulewidth
             \fi
      \ifnum0=`{\fi}}
\makeatother

\newlength{\thickarrayrulewidth}
\title{Accelerating Vision Transformer Training via a Patch Sampling Schedule}

\author {
    Bradley McDanel,
    Chi Phuong Huynh 
}
\affiliations {
    Franklin and Marshall College \\
    bmcdanel@fandm.edu, phuynh@fandm.edu
}

\begin{document}

\maketitle

\begin{abstract}
We introduce the notion of a Patch Sampling Schedule (PSS), that varies the number of Vision Transformer (ViT) patches used per batch during training. Since all patches are not equally important for most vision objectives (e.g., classification), we argue that less important patches can be used in fewer training iterations, leading to shorter training time with minimal impact on performance. Additionally, we observe that training with a PSS makes a ViT more robust to a wider patch sampling range during inference. This allows for a fine-grained, dynamic trade-off between throughput and accuracy during inference. We evaluate using PSSs on ViTs for ImageNet both trained from scratch and pre-trained using a reconstruction loss function. For the pre-trained model, we achieve a 0.26\% reduction in classification accuracy for a 31\% reduction in training time (from 25 to 17 hours) compared to using all patches each iteration. Code, model checkpoints and logs are available at \url{https://github.com/BradMcDanel/pss}.
\end{abstract}

\section{Introduction}
Vision Transformers (ViTs)~\cite{kolesnikov2021} are an adaptation of the popular Transformer model~\cite{vaswani2017attention}, conventionally applied to natural language processing (NLP), to computer vision domains. ViTs have been shown to achieve state-of-the-art performance across a wide range of tasks, such as image classification~\cite{kolesnikov2021,pmlr-v139-touvron21a}, object detection~\cite{carion2020end}, and text-to-image generation~\cite{saharia2022photorealistic}. However, while these models generally outperform other approaches, such as Convolutional Neural Networks (CNNs), they have significantly higher training and inference computational requirements due to quadratic processing time between all pairs of image patches (tokens). 

Recently, patch sparsification techniques, which select a subset of patches per image, have achieved competitive inference accuracy (i.e., with 0.5\% of the baseline ViT) using only 50\% to 70\% of the available image patches~\cite{rao2021dynamicvit,yin2022vit}. Using fewer patches per image leads to an substantial increase in images processed per second (throughput). Intuitively, this improvement can be attributed to the notion that not all patches are necessary (or even useful) for classifying a given image. However, while successfully speeding up inference, prior patch sampling approaches do not reduce the baseline ViT training time. Instead, these approaches often require additional training time compared to the baseline. This is due to (1) additional learned patch sampling modules that require access to all patches during training (even discarded ones) to learn how to sample, (2) uneven patch allocation across samples in a batch leading to parallelization inefficiencies, (3) the use of a teacher model (with access to all patches) to provide soft labels for the dynamic patch sampling student model, and/or (4) initializing a fine-tuning stage from a pretrained model that used 100\% of patches for its entire training.

In this work, we introduce a Patch Sampling Schedule (PSS) that varies the number of image patches per training iteration with the primary goal of reducing ViT training time. In a given batch, all images are allocated the same number of patches (set by a patch keep rate $\rho$). This ensure that the resulting data input to the ViT is fully dense, leading to a reduced iteration time relative to the number of kept patches.

\begin{figure}
    \includegraphics[width=\columnwidth]{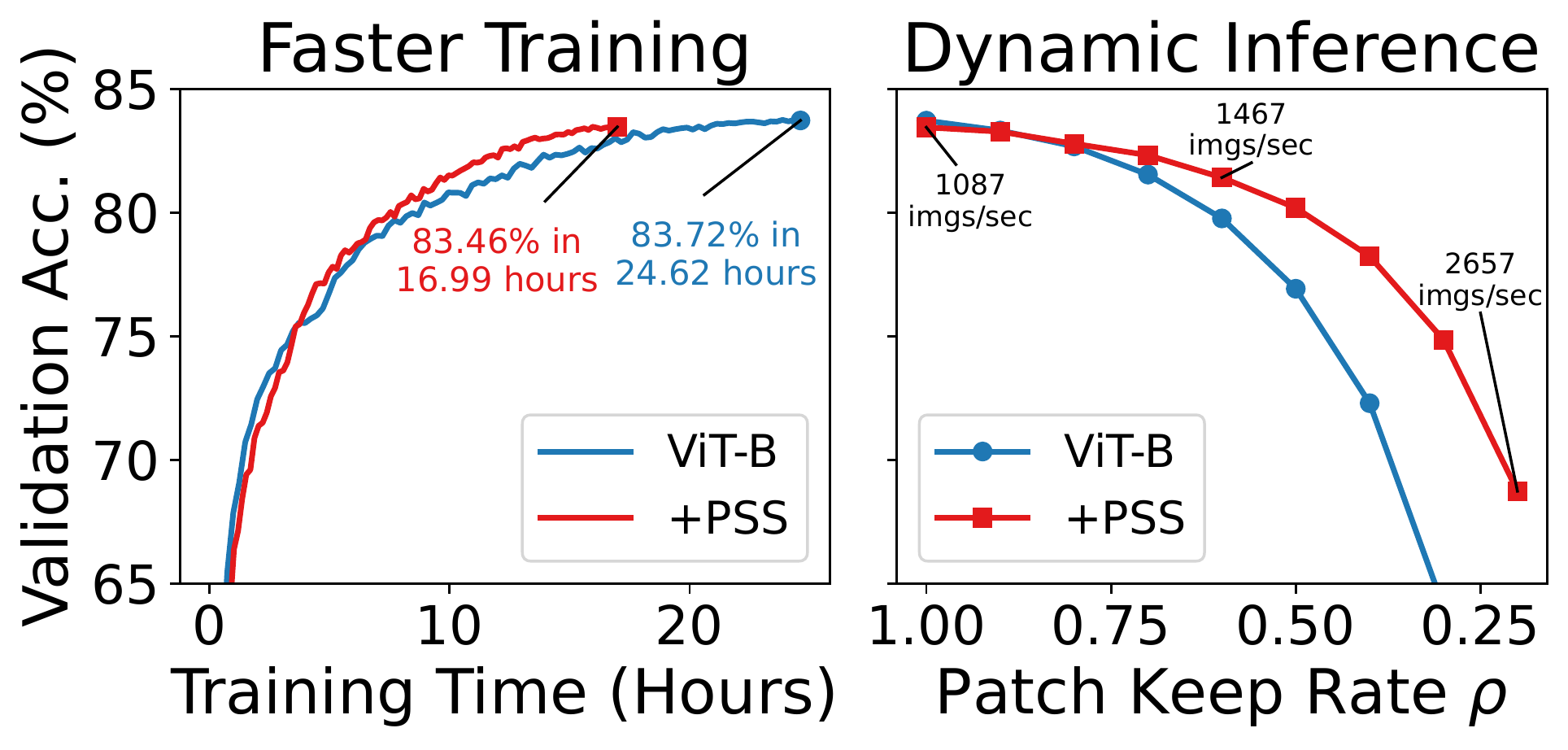}
    \caption{(left) Training ViT-B with a Patch Sampling Schedule (PSS) leads to a significant reduction in training time while achieving similar validation accuracy. The x-axis is training time (not FLOPs) for a system with 4 Nvidia A100 GPUs. (right) The ViT-B model trained with a PSS achieves a better dynamic inference results compared to the baseline as the number of kept patches per sample is varied.}
    \label{fig:overview}
\end{figure}

By progressively changing $\rho$ using a PSS, we can train substantially faster than a baseline ViT while achieving similar performance. Figure~\ref{fig:overview}(left) compares the wall clock training time for baseline ViT-B to ViT-B + PSS on ImageNet~\cite{deng2009imagenet}. The addition of the PSS leads to a training time reduction of 31\% (from 24 to 17 hours) while achieving within 0.3\% of the classification accuracy. Figure~\ref{fig:overview}(right) illustrates that the ViT model trained with a PSS is also more robust to using a subset of patches during inference. This improvement is due to the model learning from a varying number of patches across training iterations. For example, when $\rho$ = 0.6 (i.e., keeping 60\% of the patches), the ViT-B + PSS model achieves an image throughput improvement of 35\% compared to $\rho$ = 1.0 for a 2.3\% accuracy reduction. The main contributions of our work are:
\begin{itemize}
    \item A Patch Sampling Schedule (PSS) that varies the number of patches per image during training to reduce total training time. Unlike prior methods, using a PSS introduces no modifications to the ViT architecture.
    \item A higher dynamic inference accuracy curve (varying the number of patches) for models trained under a PSS than a baseline ViT using all patches each training iteration.
    \item Detailed analysis on the impact of different PSSs (e.g., linear, cyclic, etc...) and patch sampling functions (determining which patches to pick) on training time and validation accuracy.
    \item Comparison to prior work on patch sparsification for ViTs in terms of training time, inference throughput and inference accuracy for a variety of input image sizes.
\end{itemize}

\section{Related Works}
\subsection{Vision Transformers}
ViTs~\cite{parmar2018image} have recently outperformed Convolutional Neural Networks (CNNs) for various computer vision tasks, including image classification~\cite{kolesnikov2021,pmlr-v139-touvron21a}, object detection~\cite{carion2020end}, and text-to-image generation~\cite{saharia2022photorealistic}. Adapted from the Transformer models for NLP tasks, ViT tokenizes images into patches before passing them into a layer-wise attention architecture. Unlike CNNs, which use a spatial bias to look only at neighboring pixels, ViTs perform computation across all pairs of patches, leading to generally higher computational complexity. 

Before entering the ViT, each patch encodes its position in the original image stored via position embedding. In theory, this allows for permutation of patches without impacting the performance of the model. In addition to position embeddings, Transformers and ViTs may use a relative position bias~\cite{shaw2018self, liu2021swin}, which learns to encode the relative distance between each pair of patches. For patch sampling techniques, such as our proposed PSS, we find that the inclusion of the relative position bias leads to dramatic improvement in performance.


\subsection{Dynamic Inference for Vision Transformers}
Due to the high computational complexity of ViTs, there has been increased research effort on dynamically adjusting the amount of computation per sample in order to trade-off accuracy for throughput. DynamicViT~\cite{rao2021dynamicvit} introduced patch sampling during inference by adding learned control gates trained via the Gumbel-softmax technique~\cite{jang2017categorical}. These sampling modules are added at various points in the ViT and are used to select a subset of patches at each stage according to a fixed token ratio. Like our approach, DynamicViT ensures that each image have the same number of patches at a given stage in order to mitigate parallelization inefficiencies.

In contrast, A-Vit~\cite{yin2022vit} uses an Adaptive Computation Time (ACT)~\cite{graves2016adaptive} approach to dynamically sparsify the number of patches (tokens) at various stages in a ViT, leading to early exiting of tokens based on sample difficulty. While this approach may lead to better performance, as the model can adapt the token (patch) budget on a per-sample basis, the variability leads to inefficiencies in a batched setting since the input tensor now has a different number of tokens per sample. This makes ACT-based approaches, at least as currently formulated, less suitable for reducing ViT training time.

While these prior approaches focus on improving inference throughput via a patch sampling mechanism, they introduce additional overhead during training. In this work, we mainly focus on reducing ViT training time via patch sampling. To our knowledge, this is the first work to explore patch sampling with this goal. Additionally, we find the using a PSS during training naturally leads to better support for dynamic inference compared to the baseline ViT. In our evaluation, we compare our inference performance against these prior work on dynamic inference for ViTs.

\subsection{Relation to Sparse Mixture of Experts}
Sparse Mixture of Experts (MoEs)~\cite{shazeer2017outrageously} replace MLP layers in some Transformer blocks with sparse MoE layers that are partitioned across multiple devices, with each device storing a separate MLP (expert) that processes a subset of patches (tokens). Routers are used to direct subsets of patches to each expert based on their specialization. During training, \citet{riquelme2021scaling} proposed batch prioritized routing for sparse MOEs applied to ViTs to not route some patches to any expert (i.e., discard them). They show that this approach can reduce total training Floating Point Operations (FLOPs) by 20\%. While our method aims to also drop patches and increase efficiency, we do not rely on routing patches to experts. Compared to sparse MoEs, patch sampling in this work can be viewed as determining weather to route a patch to the network or to discard it entirely.

\section{Training with a Patch Sampling Schedule}
In this section, we introduce training with a patch sampling schedule (PSS), then we describe different methods for sorting and sampling patches, next we discuss different PSS schedules to employ during training, and finally, we discuss how relative position bias can be efficiently implemented when using a PSS.

\subsection{Training Overview}
\begin{figure}
    \includegraphics[width=\columnwidth]{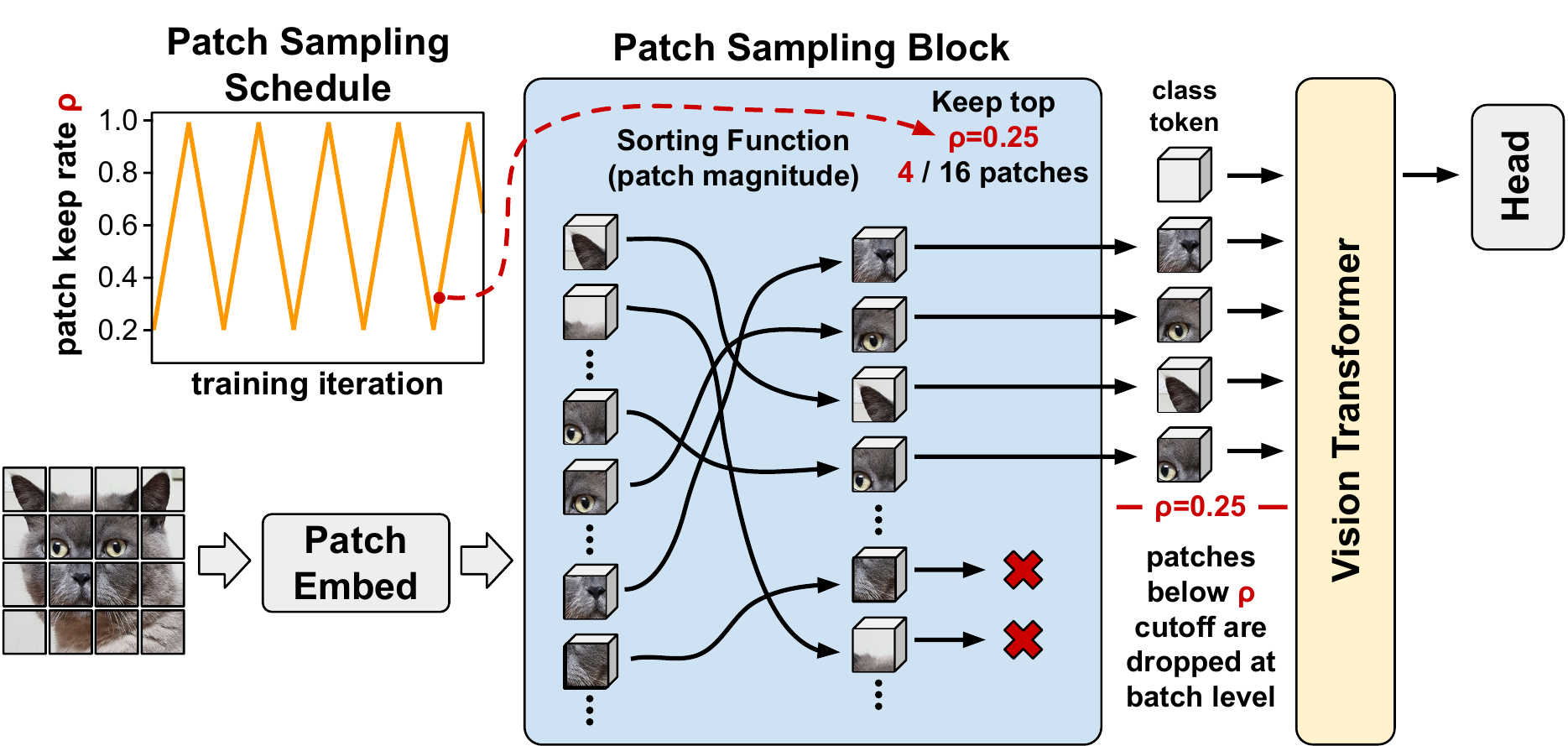}
    \caption{Overview of training a ViT with a Patch Sampling Schedule (PSS). Embedded patches are passed to a Patch Sampling Block that first orders the patches via a sorting function (e.g., patch magnitude). Then, patches below a patch keep rate $\rho$ are dropped before entering the ViT. The PSS acts like a learning rate schedule by determining $\rho$ each iteration.}
    \label{fig:fracpatch-overview}
\end{figure}

Our training framework follows the original Vision Transformer as outlined by~\citet{parmar2018image} with an augmentation that samples image patches before being passed to the ViT. Figure~\ref{fig:fracpatch-overview} provides an overview of our approach. Each image is first embedded via a patch embedding layer (a single convolution layer) that downsamples the spatial region and increases the number of channels. Then, these image patches (i.e., tokens) are passed into the patch sampling block that determines how many and which patches to keep. 

The number of patches to keep is controlled by the Patch Sampling Schedule (PSS) that varies a patch keep rate $\rho$ across training iterations. For example, in Figure~\ref{fig:fracpatch-overview}, $\rho$ is set to $0.25$ for the specific iteration, leading to only 4 of the 16 patches to be selected. The PSS varies $\rho$ across training iterations based on a set schedule that works in a similar fashion to learning rate schedules. When $\rho=1.0$, all patches are selected and training is equivalent to baseline ViT training. The specific patches that are selected are controlled by a patch sorting function $\mathcal{S}$ that orders the patches based on some criteria. In this example, the sorting function is the magnitude ($L_1$-norm) of each patch, meaning that patches with more activated embeddings are selected.

After exiting the patch sampling block, the selected patches are then passed to the ViT using an identical architecture as the baseline. Importantly, since the same number of patches are used for each image in a batch (e.g., 4 patches in Figure~\ref{fig:fracpatch-overview}), we are able to select a sub-tensor of the embedded patches that is dense. Algorithm~\ref{alg:pss-training} provides an overview of the training process using a PSS. For each training iteration, a batch of samples $X$ is first passed through the patch embedding layer, which transforms the dimensionality from $X \in \mathbb{R}^{B,C,W,H}$ to $X \in \mathbb{R}^{B,P,L}$, where $P$ is the number of patches and $L$ is the embedding dimension. Then, the patch keep rate $\rho$ is generated by the patch sampling schedule $\mathcal{P}$. Next, these embedded images are passed to the patch sampling block with $\rho$ and a patch sorting function $\mathcal{S}$ that is fixed across all training iterations. Based on the current keep rate $\rho$, fewer or more patches (from 20\% to 100\% of patches in our evaluation) are used for a given iteration.

\begin{algorithm}
  \caption{Training with Patch Sampling Schedule (PSS)\label{alg:pss-training}}
  \begin{algorithmic}[1]
    \Require{$D \in \mathbb{R}^{T,B,C,W,H}$ training images ($T$ number of batches, $B$ batch size, $C$ channels, $W$ width, $H$ height), $M$ ViT model, $e$ epochs, $\mathcal{P}$ patch sampling schedule, $\mathcal{S}$ patch sorting function}
    \Statex
    \Function{TrainWithPSS}{$D$, $M$, $\mathcal{P}$, $\mathcal{S}$}
      \For{$i \gets 0 \textrm{ to } e$}
        \For{$j \gets 0 \textrm{ to } T$}
          \Let{$X$}{$D \lbrack j \rbrack$} \Comment{$X \in \mathbb{R}^{B,C,W,H}$}
          \Let{$X$}{PatchEmbed($X$)}  \Comment{$X \in \mathbb{R}^{B,P,L}$}
          \Let{$\rho$}{$\mathcal{P}$($i*T + j$)} \Comment{$\rho$ may update each iteration}
          \Let{$\overline{X}$}{PatchSamplingBlock($X$, $\rho$, $\mathcal{S}$)}
          \Let{$Y$}{$M$($\overline{X}$)}
          \LineComment{Use $Y$ to update $M$ via backprop. Loss\\ \tabto{3.8em} function, lr, etc... are omitted for simplicity.}
        \EndFor
      \EndFor
    \EndFunction
  \end{algorithmic}
\end{algorithm}

\subsection{Patch Sampling Functions}
One of the key considerations in dynamically adjusting the number of patches across iterations is determining which patches provide the most useful information to the ViT model during both training and inference. If each patch provided the same amount of information for a given objective (e.g., classification), then random sampling would be a reasonable approach. However, as has been demonstrated by various attention-based models~\cite{hu2018squeeze,dosovitskiy2020image}, the usefulness of image regions varies widely based on their content.

Since the number of patches kept per image changes across iterations based on $\rho$, we frame sampling as first sorting the patches by a given criteria, and then selecting the top-$k$ where $k = \rho \times P$. We consider two simple patch sorting functions to use inside our patch sampling block, random sorting and magnitude sorting, which we define as:

\noindent \textbf{random sorting:} $\textrm{argsort}(i \sim \mathcal{U}(0, 1)~\textrm{where}~i \in \mathbb{R}^P)$ 

\noindent \textbf{magnitude sorting:} $\textrm{argsort}(\sum_{i=0}^L |x_i|~\textrm{where}~X \in \mathbb{R}^{P,L})$ 

For magnitude sorting, we simply use the $L_1$-norm to rank the patches. Each function uses argsort to rank the indicies of each patch. For instances, if 3 patches scored $[0.4, 0.1, 0.9]$, then their descending argsort order would be $[1, 2, 0]$.

Algorithm~\ref{alg:psb} shows how these patch sorting functions, denoted $\mathcal{S}$, are used inside a patch sampling block. Each image in a batch $X$ are passed into the sorting function $\mathcal{S}$ to get sorted patch indicies $I_X$. Then, the number of kept patches $k$ is computed by rounding to the nearest integer for the patch keep rate $\rho$. Finally, this subset of patches are selected via indexing a sub-tensor from $X$ and passed to the ViT. Note that this indexing operation is non-differentiable with respect to the indicies, and therefore the non-selected patches are not used during backpropagation.

\begin{algorithm}
  \caption{Patch Sampling Block\label{alg:psb}}
  \begin{algorithmic}[1]
    \Require{$X \in \mathbb{R}^{B,P,L}$ images after patch embed ($B$ batch size, $P$ number of patches (tokens), $L$ embed dimension), $\rho$ patch keep rate, $\mathcal{S}$ patch sorting function}
    \Ensure{$\overline{X} \in \mathbb{R}^{B,\round{\rho * P},L}$} images after patch sampling
    \Statex
    \Function{PatchSamplingBlock}{$X$, $\rho$, $\mathcal{S}$}
      \Let{$I_X$}{$\mathcal{S}(X)$} \Comment{$I_X \in \mathbb{Z}^{+B,P}$: sorted patch indicies}
      
      \Let{$k$}{$\round{\rho * P}$} \Comment{$k$: patches to keep per sample}
      
      \Let{$\overline{X}$}{$X\lbrack I_X \lbrack :, :k \rbrack \rbrack $} \Comment{$\overline{X}$: subset of patches from $I_X$}
      
      \State \Return{$\overline{X}$}
    \EndFunction
  \end{algorithmic}
\end{algorithm}

\subsection{Patch Sampling Schedules}
\begin{figure}
    \includegraphics[width=\columnwidth]{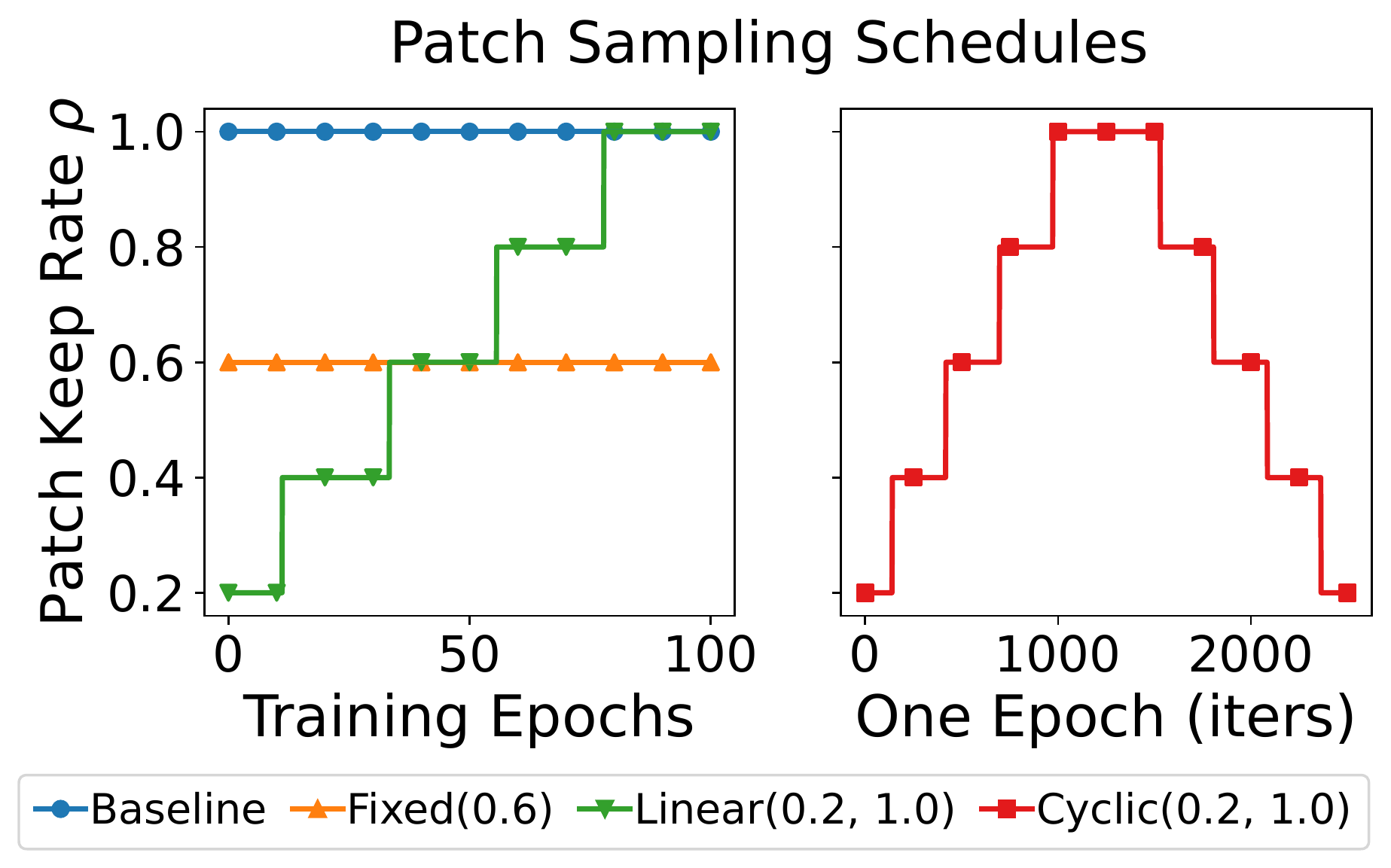}
    \caption{The patch sampling schedules used in our evaluation for 100 epochs of training. The cyclic PSS (right) repeats each epoch, so only one epoch is shown.}
    \label{fig:patch-schedules}
\end{figure}

The other main consideration for training with a PSS is the patch schedule that sets the value of $\rho$ at each training iteration. Here, we take inspiration from work on learning rate schedules and model several patch schedulers according to learning rate schedules. The schedulers using in our evaluation are depicted in Figure~\ref{fig:patch-schedules}. The baseline schedule (blue {\LARGE$\bullet$} line) uses 100\% of patches ($\rho=1.0$) for all training iterations. In our implementation, if $\rho=1.0$, then we skip the patch sampling block entirely. A Fixed PSS (orange $\blacktriangle$ line) uses the same percentage of patches (i.e., 60\%) each iteration. This is useful for comparison against other schedules that vary $\rho$ across iterations. A Linear PSS (green $\blacktriangledown$ line) linearly increases the number of patches kept across iterations. Note that we use a step function to reduce the chance of irregular dimensions (e.g., 37 patches per image) that can lead to worse running time. Finally, we use a Cyclic PSS (red {\small$\blacksquare$} line) inspired by cyclical learning rates~\cite{smith2017cyclical}. Unlike the other schedules, the Cyclic schedule has a period of a single epoch, varying $\rho$ from 0.2 to 1.0. For all schedules settings, we aim to use roughly the same amount of training time in order to fairly compare each approach.

\subsection{Relative Position Bias}
Relative position representations extend attention to consider the relative distance between tokens in an input sequence~\cite{shaw2018self}. This concept was applied to the ViT-like architecture called Swin Transformers under the name relative position bias~\cite{liu2021swin}. In practice, relative position bias is implemented using a learnable $P \times P$ matrix in each attention layer, where $P$ is the maximum number of patches. This enables additional weighting between the interaction of patches during each attention layer.

We find that the inclusion of relative position bias dramatically improves the performance of training with a PSS. However, in order to implement relative position bias efficiently, we must index the $P \times P$ relative position bias matrix for each image in a batch in order to add the specific biases for each image. This is because our patch sorting function $\mathcal{S}$ does not have the restriction that all images must select the same patches across a batch. This per-image indexing adds minor overhead to our method, which we include in all reported training and inference timings.

\section{Evaluation}
We evaluate using PSSs on ViTs for ImageNet~\cite{deng2009imagenet} both trained from scratch and pre-trained using a reconstruction loss function. To train ViT from scratch, we modify the codebase (\url{https://github.com/facebookresearch/deit}) of DeiT~\cite{Touvron2022DeiTIR} to be compatible with PSSs. Additionally, we add relative position bias to each attention layer in DeiT starting from the implementation in the SimMIM~\cite{xie2022simmim} codebase (\url{https://github.com/microsoft/SimMIM}). For pre-trained models, we modify the SimMIM codebase to include PSSs and start fine-tuning using their models trained using reconstructive loss. Otherwise, we follow the training regime and data augmentation used by each project for their baseline results. We train all models using a single machine with 4 A100 GPUs. 

\subsection{Reduction in Training Time}
\begin{figure}
    \includegraphics[width=\columnwidth]{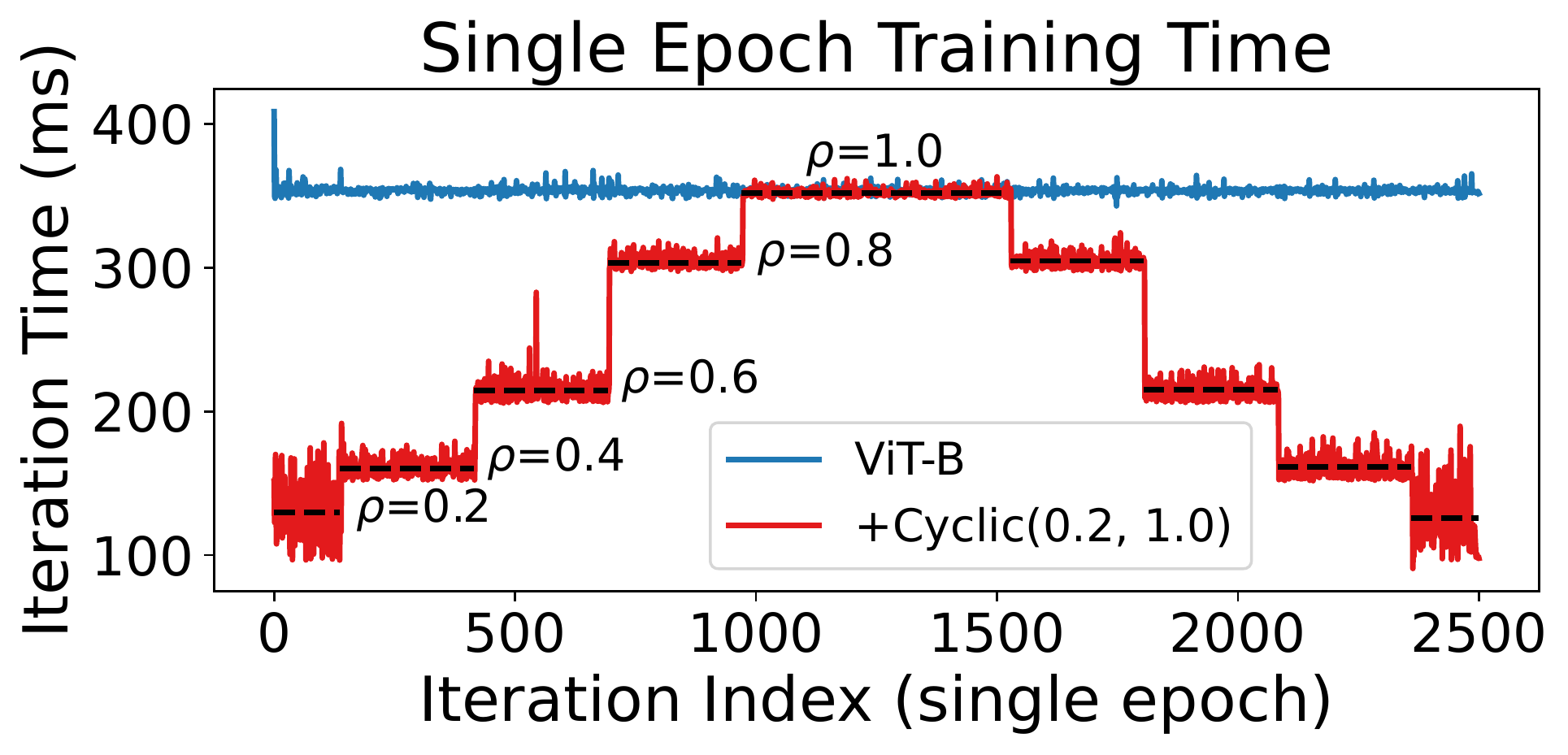}
    \caption{Comparing per-iteration training time for ViT-B and ViT-B + Cyclic(0.2, 1.0) patch sampling schedule. The Cyclic PSS leads to a 31\% reduction in training time.}
    \label{fig:train-times}
\end{figure}

Using a PSS, we can change runtime per training iteration by varying the patch keep rate $\rho$. Figure~\ref{fig:train-times} shows the iteration times across a single epoch for the baseline (ViT-B) and ViT-B with a Cyclic PSS with $\rho$ varying from $0.2$ to $1.0$ using the cyclic patch schedule shown in Figure~\ref{fig:patch-schedules}. Using 4 A100 GPUs, this leads to a per-epoch training time on ImageNet of 610 seconds (10.1 minutes) when using the cyclic PSS compared to 890 seconds (14.8 minutes) for the baseline.

Due to overhead associated with indexing the correct relative position biases for patches selected in each image, using PSS with $\rho$ close to 1 leads to worse iteration time than the baseline. Therefore, for $\rho=1.0$, we skip the patch sampling block entirely and train in the same fashion as the baseline approach. In Figure~\ref{fig:train-times} this is reflected for the region where $\rho=1.0$, as both approaches have approximately the same iteration time in this region.

\subsection{Impact of PSF and PSS on Performance}
In this section, we compare the training time and ImageNet top-1 validation accuracy for combinations of patch sorting functions (PSFs) and patch sampling schedules (PSSs). Table \ref{tab:psf-pss} reports accuracy various PSF/PSS combinations on the pre-trained ViT-B model fine-tuned on ImageNet. Generally, using a PSS substantially reduces training time with only a small loss in accuracy compared to the baseline. Specifically, our Cyclic PSS achieves 31\% reduction in training time with a 0.26\% loss in final validation accuracy.

Comparing the results of the two patch sorting functions, we observed that magnitude-based sorting achieves significantly higher accuracy regardless of the PSS used. This suggests that the magnitude of patch embeddings is a simple but useful metric for determining how to sort the patches in terms of importance. Additionally, we find that a fixed PSS that always selects $\rho=0.6$ (60\% of patches) performs significantly worse than the other approaches. This makes intuitive sense, as the model trained with $\rho=0.6$ will not be aligned to using 100\% of the patches during inference. Finally, we see that the Cyclic PSS slightly outperforms the Linear PSS, using either random or magnitude PSF. Based on this, we use the Cyclic PSS when comparing to other approaches.


\begin{table}
\renewcommand{\arraystretch}{1.4}%
\begin{adjustbox}{width=\columnwidth, center}
\begin{tabular}{l|lll}
\thickhline
PSF                        & PSS              & Training Time $\downarrow$ &   Top-1 Accuracy $\uparrow$\\ \thickhline
Baseline                   & Baseline         & 24.6h              & 83.72\%                         \\ \hline
\multirow{3}{*}{Random}    & Fixed(0.6)       & 15.1h \cc{(-38.6\%)}    & 82.76\% \cc{(-0.96\%)}              \\ 
                           & Linear(0.2, 1.0) & 17.0h \cc{(-31.1\%)}    &  83.24\% \cc{(-0.48\%)}                       \\
                           & Cyclic(0.2, 1.0) & 17.0h \cc{(-31.0\%)}    &  83.29\%  \cc{(-0.43\%)}                      \\ \cline{1-2} \cline{3-4} 
\multirow{3}{*}{Magnitude} & Fixed(0.6)       & 15.1h \cc{(-38.6\%)}    & 83.09\% \cc{(-0.63\%)}                       \\  
                           & Linear(0.2, 1.0) & 17.0h \cc{(-31.0\%)}    & 83.27\% \cc{(-0.45\%)}              \\ 
                           & Cyclic(0.2, 1.0) &  17.0h \cc{(-31.0\%)}   & 83.46\% \cc{(-0.26\%)}                         \\
\thickhline
\end{tabular}
\end{adjustbox}
\caption{The training time and ImageNet top-1 validation accuracy for combinations of patch sorting functions (PSFs) and patch sampling schedules (PSSs) on pre-trained ViT-B. $\rho=1.0$ when computing top-1 accuracy for all settings.}
\label{tab:psf-pss}
\end{table}

\subsection{Dynamic Inference by Varying Patch Keep Rate}
\begin{figure}
    \includegraphics[width=\columnwidth]{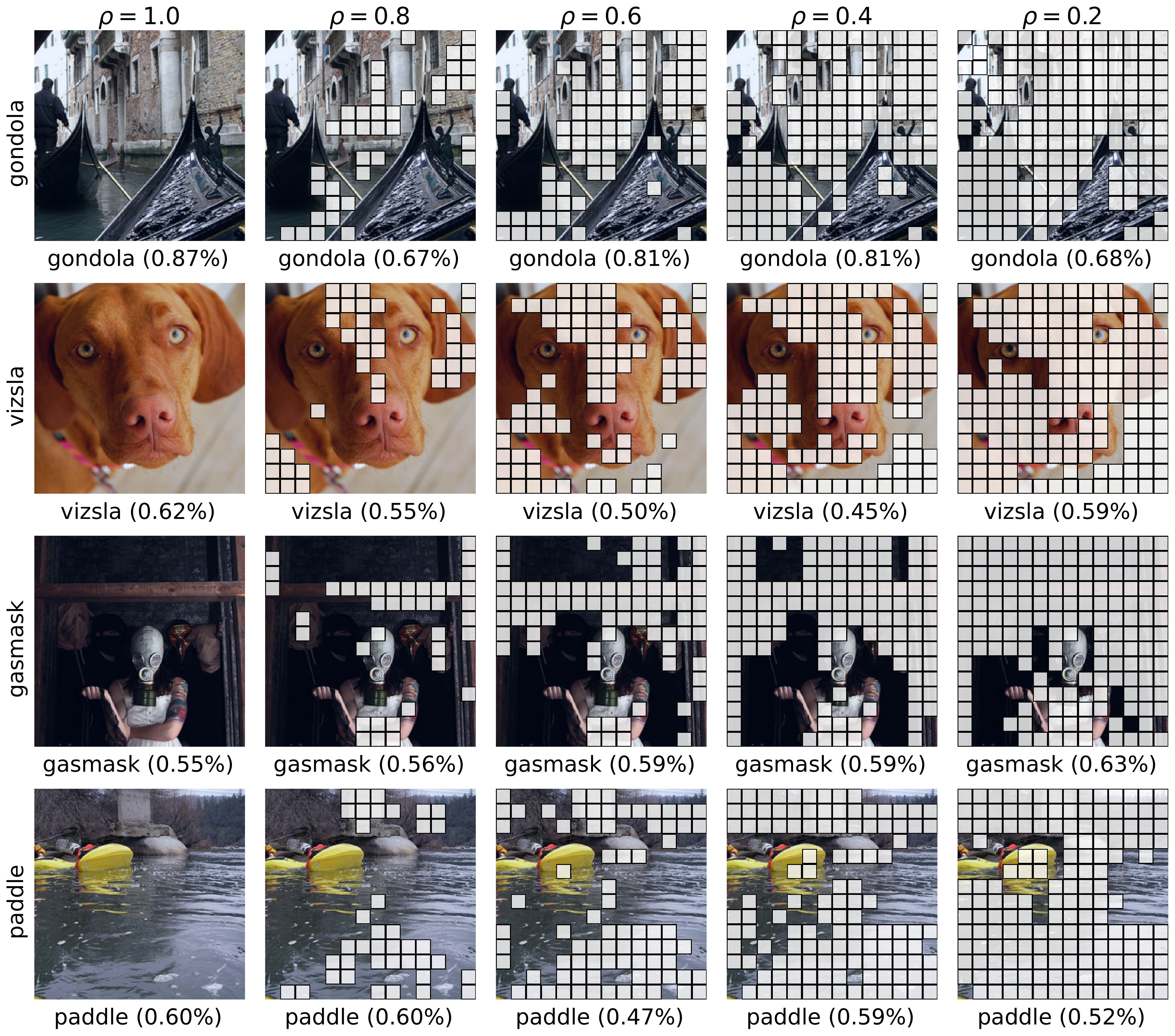}
    \caption{Visualization of kept patches in four images for different $\rho$ values using the ViT-B model trained with magnitude PSF and cyclic PSS. Each row label is the ground-truth, and each label below an image is the prediction with top-1 probability in parentheses.}
    \label{fig:image-patches}
\end{figure}

Prior work on ViT patch sparsification target a fixed patch keep rate during training (e.g., $\rho=0.7$) with the goal of minimizing accuracy loss for this single setting. By comparison, models trained with a PSS are able gradually trade-off accuracy for throughput by varying $\rho$. Figure~\ref{fig:image-patches} shows the kept patches for a batch of four images under PSS using magnitude sorting as $\rho$ is varied from $1.0$ to $0.2$. For a given $\rho$ value, all images within the batch use the same number of patches. For instance, when $\rho=0.2$, each images retains exactly 39 patches (38 of the 14$\times$14 = 196 patches and the single class token). We observe that simply using the magnitude of patch embeddings as a metric to determine patch importance leads to observably meaningful results. For instance, for the second row in Figure~\ref{fig:image-patches} of the ``vizsla'', we see that the kept patches focus on defining characteristics of the object class, such as its nose, ears, and eyes. At $\rho=0.2$, the majority of unrelated background patches have been dropped from the patch sampling block.

\begin{figure}
    \includegraphics[width=\columnwidth]{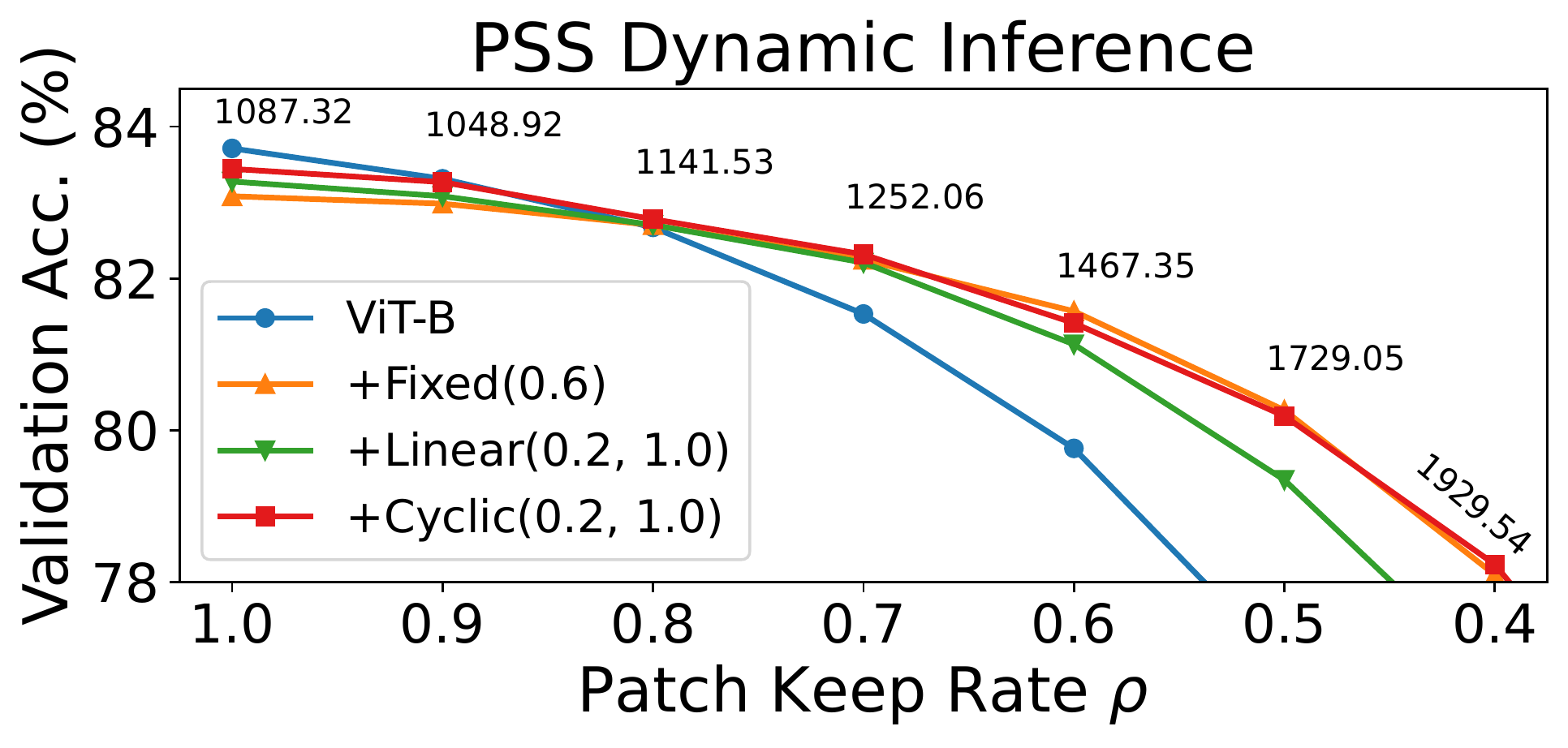}
    \caption{Comparing the dynamic inference performance of baseline ViT-B to ViT-B trained using proposed PSSs by varying the patch keep rate $\rho$. All three schedules have a similar training time. Images are processed in batches of 256. Throughput (images/sec) are shown above each $\rho$ setting.}
    \label{fig:sweep-drop-ratio}
\end{figure}

Figure~\ref{fig:sweep-drop-ratio} shows the dynamic inference performance of the ViT-B (baseline) model with the PSSs shown in Table~\ref{tab:psf-pss} using the magnitude PSF. For the baseline model, we injected the patch sampling block into the model graph after training in order see how it would perform under different $\rho$ settings. While the baseline approach has higher accuracy when $\rho=1.0$, the validation accuracy drops sharply as $\rho$ approaches $0.5$ compared to the PSS settings. This is likely due to the baseline model being trained with 100\% of patches in each training iteration and therefore not being accustomed to dealing with substantially fewer patches.

The Fixed(0.6) PSS schedule has the lowest performance at $\rho=1.0$ to $\rho=0.8$, which is also likely due to it only seeing 60\% of patches during each training iterations. By comparison, the linear PSS performs similar to the cyclic PSS for the $\rho=1.0$ to $\rho=0.8$ range, but does significantly worse in the $\rho=0.6$ to $\rho=0.4$ range. While the linear PSS did get get some training iterations where  $\rho=0.4$, it was at an early stage of training (refer back to Figure~\ref{fig:patch-schedules}) and likely the weights slowly adapted away from this region in the later stages of training. Finally, the cyclic PSS appears to provide the best dynamic range overall, by doing similar to linear in the  $\rho=1.0$ to $\rho=0.8$ range and similar to fixed(0.6) in the $\rho=0.6$ to $\rho=0.4$ range.

To our knowledge, this is the first approach which enables this type of dynamic inference by simply changing the number of patches kept in a patch sampling block during inference. As we discuss in the next section, prior work on patch sparsification, DynamicVit~\cite{rao2021dynamicvit} and A-ViT~\cite{yin2022vit}, aim to find a fixed sparsity setting (e.g., $\rho=0.7$) that achieves a minimal reduction in accuracy while leading to faster inference. By training with a PSS, we can achieve a dynamic inference range, leading to a natural trade-off between inference accuracy and runtime. These types of dynamic approaches may be useful in energy-constrained settings (e.g., mobile devices) where this type of trade-off can be made based on current battery conditions~\cite{cai2019once}.

\subsection{Comparison to Patch Sparsification Techniques}
Prior work has focused on improving inference throughput of ViT models via patch sparsification. However, these approaches lead to an increased training time compared to the baseline models for multiple reasons. First, these approaches apply their token sparsification approach as a fine-tuning step after the initial model (e.g., DeiT-S) has already been trained on ImageNet using 100\% of patches for 300 epochs. Therefore, any additional training to achieve better results using fewer patches can be considered training overhead compared to the baseline approach. Second, these approaches often use a teacher model (i.e., distillation) to provide more accurate soft labels (or use an additional distillation token as in DeiT) to aid in learning the token sparsification approach. These teacher models are usually the original pretrained model that these approaches are augmenting to do token sparsification. Therefore, the use of a teacher model dramatically increases the training iteration time during fine-tuning stage. Third, the techniques to learn which patches to drop increase training time, as additionally parameters/layers related to determining which patches to keep are learned during the fine-tuning stage. This requires dropped patches to be kept around for updating the weights during backpropagation. For instance, DynamicViT uses the Gumbel-softmax technique~\cite{jang2017categorical} to learn a hard threshold function during training.

These approaches also require additional hyperparameters tied to loss functions that encourage token sparsification. While additional loss functions are common in many applications, they often require parameter search for the associated hyperparamters to tune the level of impact of the loss function (e.g., amount of token sparsification applied). These learnable components may also have some impact on regularization. For instance, A-Vit~\cite{yin2022vit} disables MixUp regularization~\cite{zhang2017mixup} during training as they note it is not compatible with their approach. We observe that our simpler patch selection approach (with no learnable component) performs best with all regularization enabled as is done by the baseline model. Likely, pixel mixing across the same patch location makes it significantly more challenging to learn a patch selection layer.

Table~\ref{tab:other-approaches} provides a detailed training and inference comparison for the DeiT-S baseline, Dynamic-Vit, A-ViT, and our PSS approach using a magnitude PSF and cyclic PSS. The table is divided into two sections based on a $224 \times 224$ or $384 \times 384$ image size. Models trained with a teacher model during the fine-tuning epochs are denoted with a \cmark.

For the $384 \times 384$ image size, we observe that our PSS approach is able to achieve comparable or better inference performance to other methods while significantly reducing the training time. For instance, our PSS approach reduces the training time by 26.9\% while achieving 1.3\% higher validation accuracy than DynamicViT. Our model actually outperforms the baseline DeiT-S model reported in DynamicVit by 1.1\%. We believe this is mostly due to our inclusion of relative position bias when training models with PSS (discussed in next section). Additionally, our model achieves a 26\% reduction in training time on a 4 A100 GPU system compared to training the baseline model on the same system. Compared to DynamicVit, we are able to achieve a 36.2\% reduction in training time, as we do not require additional finetuning epochs or the use of a teacher model. Finally, our model can dynamically trade-off throughput for accuracy by changing $\rho$ as demonstrated by the 3 rows with $\rho=1.0$, $\rho=0.7$, and $\rho=0.5$. For $\rho=0.5$, our model doubles the throughput of the baseline while losing only 0.6\% accuracy.

For the $224 \times 224$ image size, using a PSS was unable to achieve the same dramatic improvement in training time reduction. Our system, with an AMD EPYC 7313 16-Core Processor, became CPU bottlenecked performing image decompression for DeiT-S for smaller patch keep rates (e.g., $\rho=0.2$). Still, we reduced training time by 7.6\% compared to the baseline while achieve similar performance. Note that in this case, we trained the baseline DeiT-S model achieving a 80.1\% accuracy. When computing throughput, we use a single A100 and can therefore devote all CPU cores to that single GPU. Under this setting, we see this inference setting, we observe that our model can dramatically increase throughput (by 72.8\% for $\rho$ = 0.5). Thus, given a more powerful CPU, we expect to see improved training time for smaller image resolutions.

A-ViT fine-tunes for 100 additional epochs compared to only 30 for DynamicVit. This leads A-ViT to have a significantly longer total training time compared to the other approaches. 

\begin{table*}
\renewcommand{\arraystretch}{1.4}%
\centering
\begin{adjustbox}{width=\textwidth, center}
\begin{tabular}{lccccc|ccccc}
\thickhline
Model &  \makecell{Image\\Size} & \makecell{Pretrain\\Epochs $\downarrow$} & \makecell{Finetune\\Epochs $\downarrow$} & \makecell{Teacher\\Model} & \makecell{Training\\Time $\downarrow$} & {\LARGE$\rho$} & \makecell{Inference\\GFLOPs $\downarrow$} & \makecell{Throughput\\(images/second) $\uparrow$} & \makecell{Top-1\\Accuracy $\uparrow$} \\ \thickhline
DeiT-S (baseline) & 224 & 300 & 0 & \xmark & 29.9h & 1.0 & 4.6 & 1592 & 80.1\% \\
DynamicViT-S & 224 & 300 & 30 & \cmark & 34.9h$^\dagger$ \cc{(+14.3\%)} & 0.7 & 2.9 & 2191 \cc{(+37.6\%)}& 79.3\%$^*$ \cc{(-0.8\%)} \\ 
A-ViT-S & 224 & 300 & 100 & \xmark & 44.2h$^\dagger$ \cc{(+47.8\%)} & N/A & 3.6$^\mathsection$ & N/A$^\ddag$ & 78.6\%$^\mathsection$ \cc{(-1.5\%)} \\ 
A-ViT-S & 224 & 300 & 100 & \cmark & N/A$^\ddag$ & N/A & 3.6$^\mathsection$ & N/A$^\ddag$ & 80.7\%$^\mathsection$ \cc{(+0.6\%)} \\ \hline
\multirow{3}{*}{DeiT-S+PSS (ours)} & \multirow{3}{*}{224} & \multirow{3}{*}{300}& \multirow{3}{*}{0} & \multirow{3}{*}{\xmark} & \multirow{3}{*}{27.9h \cc{(-7.6\%)}} & 1.0 & 4.6 & 1596 \cc{(+0.2\%)} & 80.4\% \cc{(+0.3\%)} \\
&  &  &  & &  & 0.7 & 3.2 & 2233 \cc{(+40.2\%)} & 79.4\% \cc{(-0.7\%)} \\
&  &  &  & &  & 0.5 & 2.2 & 2754 \cc{(+72.8\%)} & 77.5\% \cc{(-2.6\%)} \\ \thickhline
DeiT-S (baseline) & 384 & 300 & 0 & \xmark & 149.2h$^\dagger$ & 1.0 & 15.5 & 471 & 81.6\%$^*$  \\
DynamicViT-S & 384 & 300 & 30 & \cmark & 170.9h$^\dagger$ \cc{(+14.5\%)} & 0.7 & 9.5 & 637 \cc{(+35.2\%)} & 81.4\%$^*$ \cc{(-0.2\%)} \\ \hline
\multirow{3}{*}{DeiT-S+PSS (ours)} & \multirow{3}{*}{384} & \multirow{3}{*}{300} & \multirow{3}{*}{0} & \multirow{3}{*}{\xmark} & \multirow{3}{*}{109.0h \cc{(-26.9\%)}} & 1.0 & 15.5 & 465 \cc{(-1.2\%)} & 82.7\% \cc{(+1.1\%)}\\
&  &  &  &  &  & 0.7 & 10.3 & 628 \cc{(+33.3\%)} & 82.1\% \cc{(+0.5\%)} \\
&  &  &  &  &  & 0.5 & 7.1 & 956 \cc{(+103.3\%)} & 81.0\% \cc{(-0.6\%)}\\
\thickhline
\end{tabular}
\end{adjustbox}
\caption{Training and inference results for patch sparsification techniques applied to DeiT-S for two image resolutions (224$\times$224 and 384$\times$384). Throughput is computed on a single A100 GPU using all CPU cores for image decompression. All models have 22M parameters. $^{*\mathsection}$Reported numbers taken from prior work (training not reproduced). $^*$DynamicVit~\cite{rao2021dynamicvit}. $^{\mathsection}$A-ViT~\cite{yin2022vit}. $^\dagger$Estimated from single epoch using 4 A100 GPUs for GitHub repositories of DynamicViT and A-Vit. The pre-training time (e.g., DeiT-S 224$\times$224) is included in the total training time. $^\ddag$Model and/or code not provided by related work, so we could not benchmark training time or throughput.}
\label{tab:other-approaches}
\end{table*}

\subsection{Impact of Relative Position Bias}
We find that the inclusion of relative position bias~\cite{liu2021swin} in each attention layer has a significant impact on the final top-1 accuracy of ViT models trained with PSS. Table~\ref{tab:rel-pos} shows the accuracy of DeiT-S trained with and without relative position bias. When relative position bias is added to the baseline DeiT-S, it leads to a 1\% improvement in accuracy (from 80.1\% to 81.1\%). By comparison, the DeiT-S+PSS model without relative position bias achieves a 1.9\% lower accuracy than the baseline. When relative position bias is included with PSS, it outperforms the baseline without it by 0.3\%. We hypothesize that the use of relative position bias becomes more important to attention when fewer patches are observed during many training iterations. 

\begin{table}
\renewcommand{\arraystretch}{1.4}%
\centering
\begin{adjustbox}{width=0.9\columnwidth, center}
\begin{tabular}{lcccc}
\thickhline
Model &  \makecell{Image\\Size} & \makecell{Relative\\Position Bias} & \makecell{Top-1\\Accuracy $\uparrow$} \\ \thickhline
DeiT-S (baseline) & 224 & \xmark & 80.1\% \\
DeiT-S+PSS (ours) & 224 & \xmark & 78.2\% \cc{(-1.9\%)}  \\ 
DeiT-S & 224 & \cmark & 81.1\% \cc{(+1.0\%)} \\
DeiT-S+PSS (ours) & 224 & \cmark & 80.4\% \cc{(+0.3\%)} \\
\thickhline
\end{tabular}
\end{adjustbox}
\caption{Impact on accuracy of adding relative position embedded to each attention layer for DeiT-S. $\rho=1.0$ when computing top-1 accuracy for all settings.}
\label{tab:rel-pos}
\end{table}

\section{Future Directions}
\noindent \textbf{Adaptation to Other Domains:} While we focused exclusively on ViT models in this work, there is reason to think the proposed patch sampling approach could reduce training costs for other Transformer models. For instance, CALM~\cite{schuster2022confident} is a recent dynamic inference approach for generative language modeling which uses dynamic depth per generation in a similar fashion to A-ViT. Another promising domain is text-to-image synthesis, where different image regions likely require different amount of computation for pixel generation.

\noindent \textbf{Multiple Stages of Patch Sampling:} Both Dynamic-ViT and A-ViT use multiple patch sampling stages that progressively sparsify samples as they get deeper in the network. From some brief experiment, we found little difference in final accuracy for PSS when using one or more stages. However, we did not have the computational resources to perform extensive grid search for multi-stage sampling.

\noindent \textbf{Efficiently Learning Sampling Model:} Our magnitude-based patch sorting functions is a simple but reasonable effective approach for scoring patches. However, it may be possible to further improve performance by learning a patch sampling sub-network. The challenge is how such a sampling network can be learned without dramatically increasing the training time as we observed with the prior work. 

\section{Conclusion}
In this paper, we propose a Patch Sampling Schedule (PSS), that varies the number of patches per iteration of ViT training. We find that our approach dramatically reduces training time of larger models by 20-30\% compared to the baseline with leading to similar classification accuracy (within 0.3\%). Additionally, training with a PSS allows the model to achieve improved accuracy across a wider dynamic inference range (by varying $\rho$). Compared to prior work that focus only on patch sparsification during inference, we are able to achieve similar accuracy for a given sparsity setting. Additionally, our approach reduces training time instead of increasing it like prior approaches. We hope that our simple patch sampling mechanism applied to ViT models will encourage others to look into the effectiveness of patch sampling as a means of reducing training time for other Transformer domains (e.g., object detection, generative language modeling, and text-to-image synthesis).

\bibliography{aaai23.bib}
\newpage

\section{Appendix}
\subsection{Evaluating PSS over Training}
In this section, we show the training performance of ViT-B models pretrained on an image reconstruction loss taken from SimMIM~\cite{xie2022simmim} and DeiT-S~\cite{Touvron2022DeiTIR} models trained from scratch on ImageNet~\cite{deng2009imagenet}. We use the magnitude patch sorting function and cyclic patch sampling schedule for all PSS settings. Since SimMIM only requires 100 epochs of finetuning (due to pretraining), the ViT-B settings finish substantially faster than the DeiT ones. This faster training time is why we chose to use the ViT-B model to evaluate the impact of PSF and PSS combinations in our main evaluation. All DeiT models were trained for 300 epochs to match the DeiT experimental setup. The 224 or 384 in each model name denote the image size input to the model during training (e.g., DeiT-S-224 uses 224$\times$224 and DeiT-S-384 uses 384$\times$384). 

\begin{figure}[ht]
    \includegraphics[width=\columnwidth]{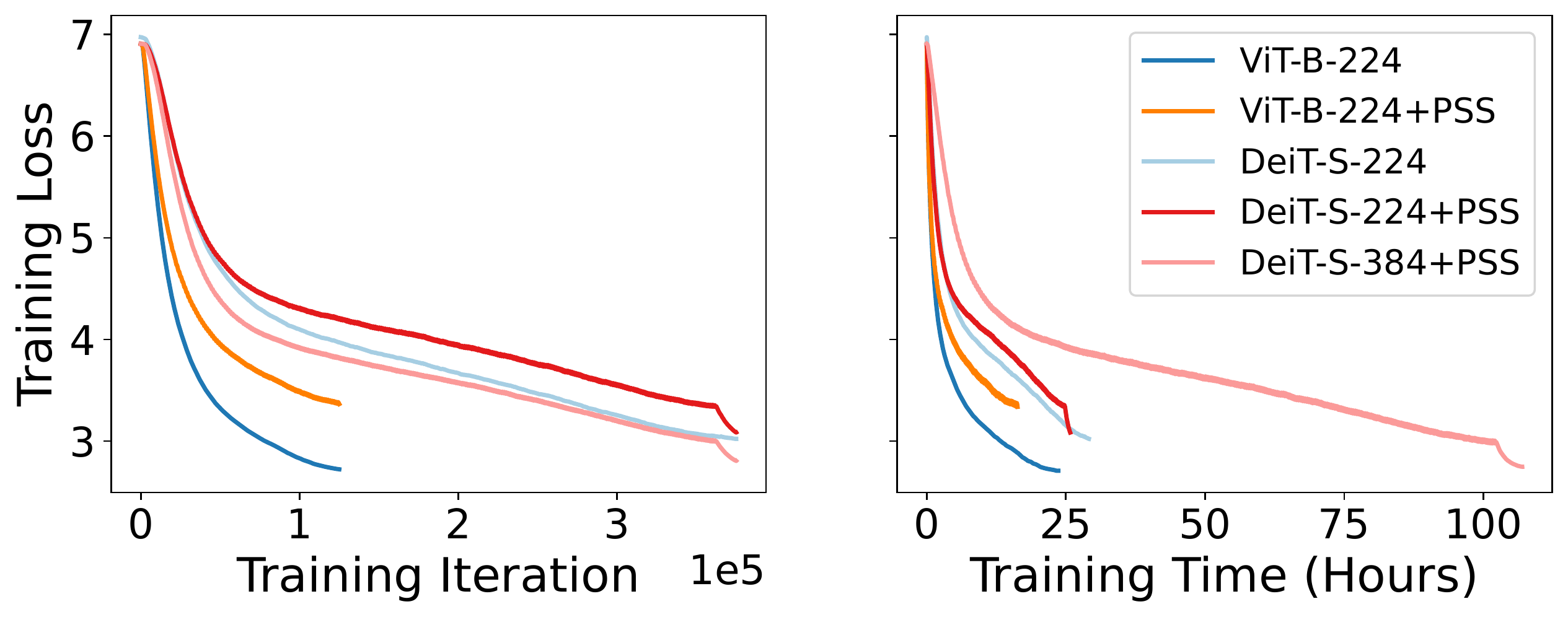}
    \caption{Training loss for pretrained ViT-B (SimMIM) and DeiT models. (left) x-axis is training iteration. (right) x-axis is training time on 4 A100 GPU machine. All curves are smoothed via an exponential moving average.}
    \label{fig:training-loss}
\end{figure}

Figure~\ref{fig:training-loss} shows the training loss for all settings. The left sub-figure shows the training loss across number of training iterations, while the right sub-figure shows the training loss across training time in hours on a 4 A100 GPU machine. As reported in the main evaluation, the ViT-B+PSS achieves a 31\% reduction in training time while the DeiT-S-224+PSS model achieves only a 7\% reduction in training time.

\begin{figure}[ht]
    \includegraphics[width=\columnwidth]{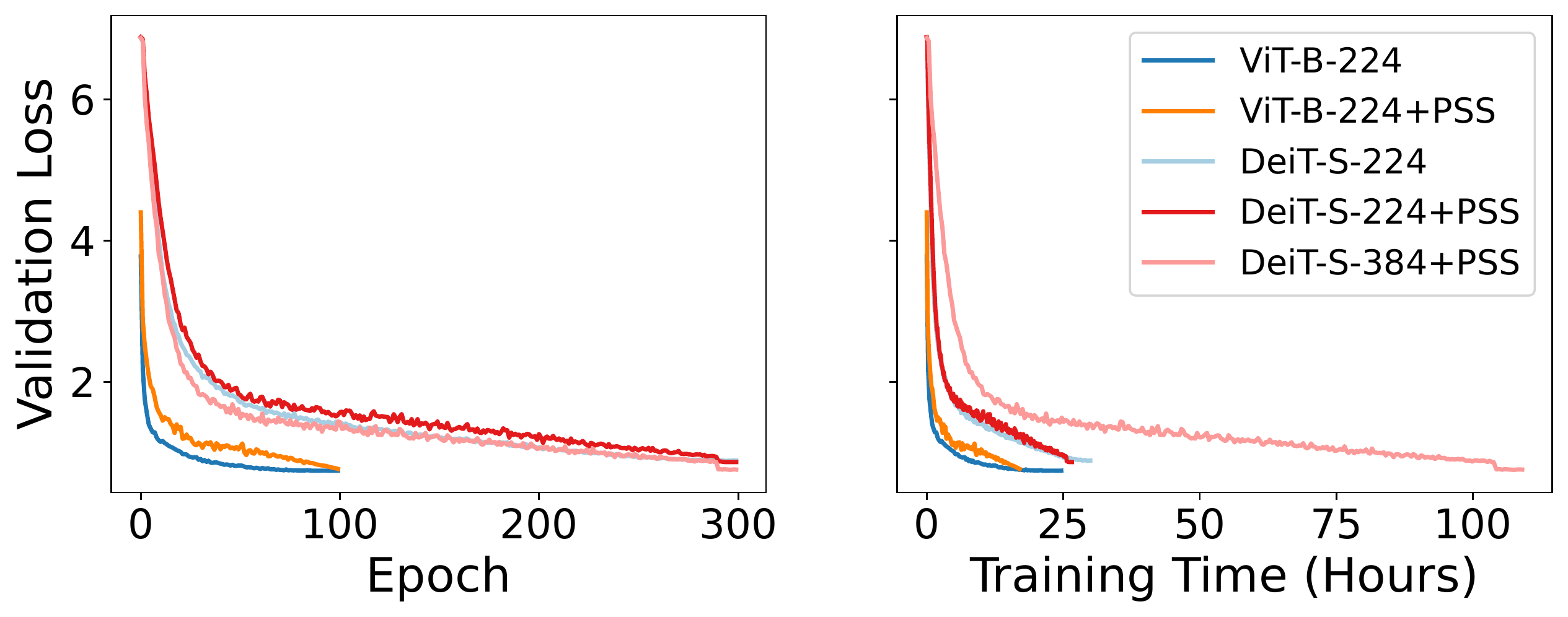}
    \caption{Validation loss for pretrained ViT-B (SimMIM) and DeiT models. (left) x-axis is validation epoch. (right) x-axis is validation loss over training time (hours).}
    \label{fig:validation-loss}
\end{figure}

\begin{figure}[ht]
    \includegraphics[width=\columnwidth]{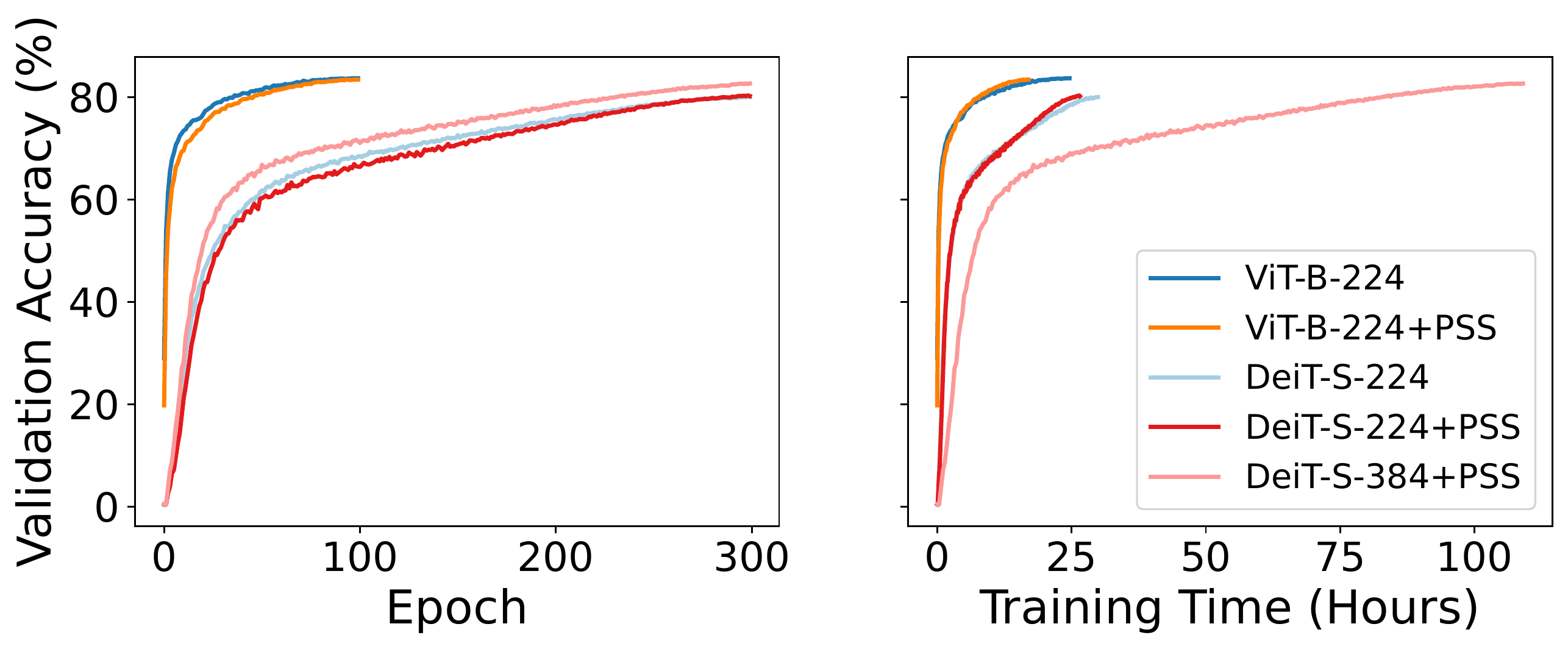}
    \caption{Top-1 validation accuracy (\%) for pretrained ViT-B (SimMIM) and DeiT models. (left) x-axis is validation epoch. (right) x-axis is validation loss over training time (hours).}
    \label{fig:validation-accuracy}
\end{figure}

Figure~\ref{fig:validation-loss} shows the validation accuracy for all models in terms of epochs (left) and training time (right). We observe similar a similar trend as in the training loss figure. Note that the models with PSS converge to roughly the same validation loss as their baseline models. Figure~\ref{fig:validation-accuracy} shows the top-1 validation accuracy over the course of training. Again, the left sub-figure shows accuracy improvement over epochs while the right sub-figure shows accuracy improvement over training time.

\subsection{Spatial Position of Kept Patches}
\begin{figure*}
    \includegraphics[width=\textwidth]{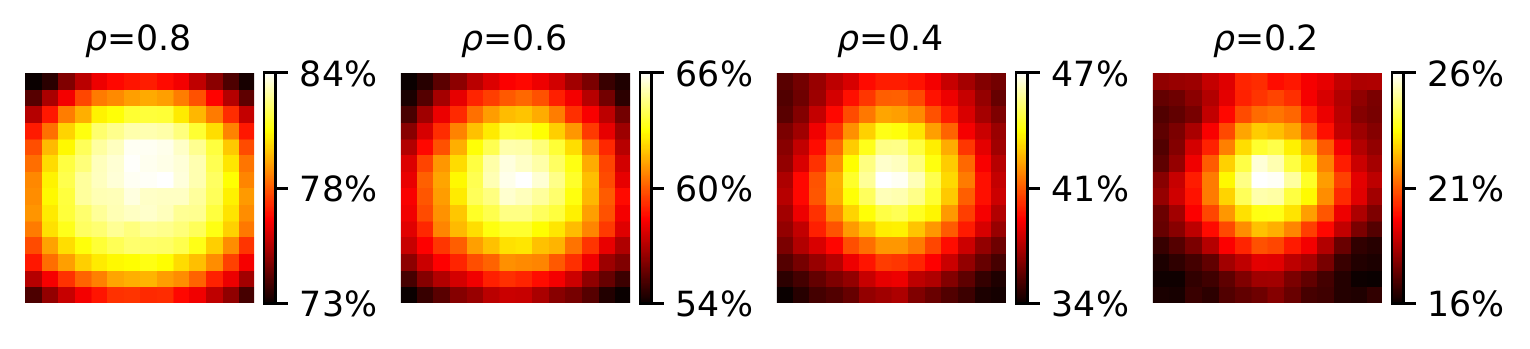}
    \caption{Distribution of the spatial position of kept patches in ImageNet validation set as $\rho$ is varied from 0.8 to 0.2. When $\rho=1.0$ all patches in each spatial position are kept. This uses the ViT-B model with Magnitude PSF and Cyclic PSS.}
    \label{fig:patch-distribution}
\end{figure*}

Figure~\ref{fig:patch-distribution} shows the distribution of which patches are kept more frequently across spatial positions during patch sampling. When $\rho=1.0$, all patches are kept, so there is no useful information to show. For a given $\rho$ (e.g., $\rho=0.8$), we observe that the difference between the most frequently and least frequently kept patch is only around 10\%. This suggests that there is reasonably large spatial variance across images, making it difficult to use a simpler sampling approach. On average, we see that patches are sampled more often from the center of the image. This is likely due to the classification target frequently appearing in the center of ImageNet images.

\subsection{DynamicViT Dynamic Inference Performance}
\begin{figure}[ht]
    \includegraphics[width=\columnwidth]{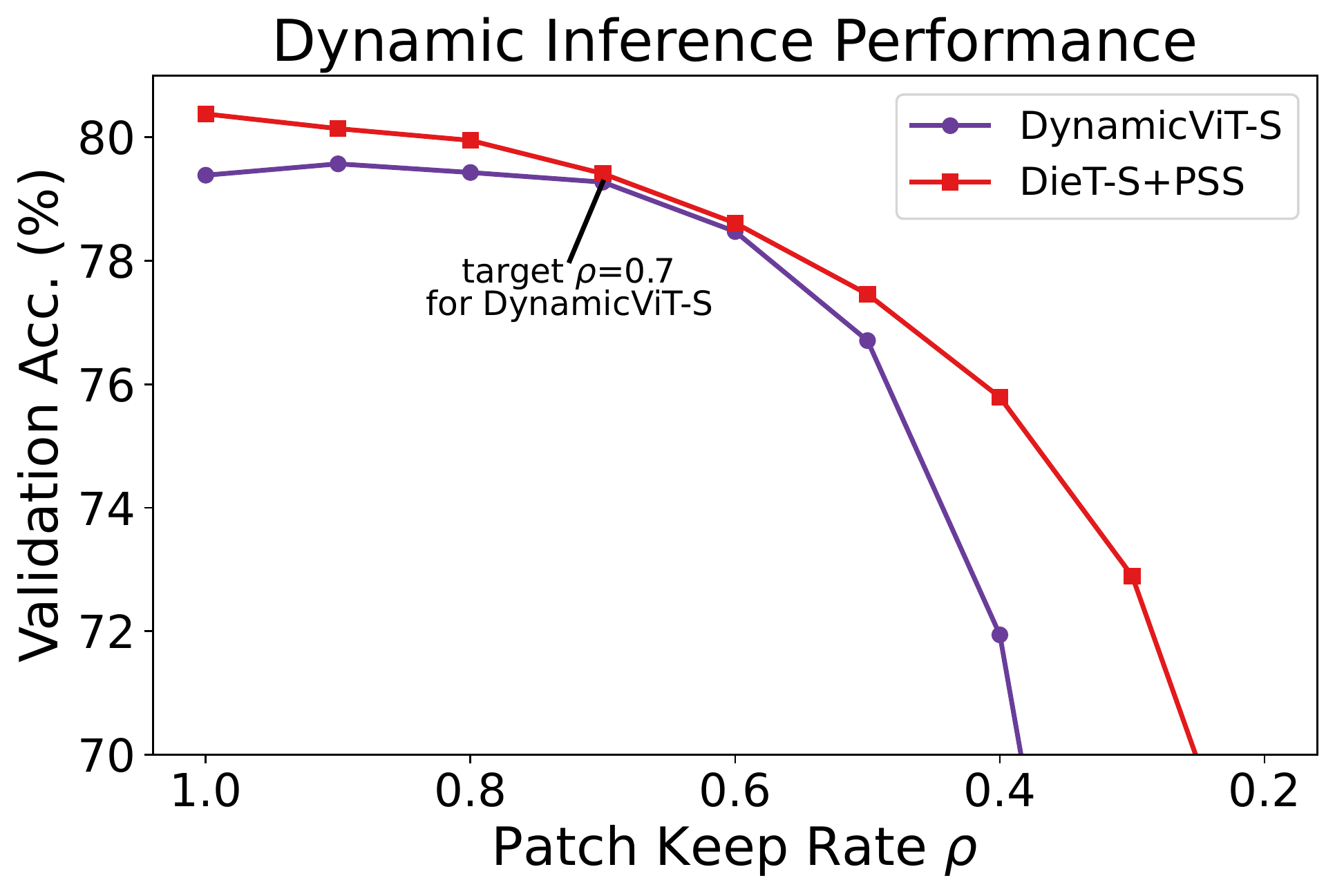}
    \caption{Varying the patch keep rate $\rho$ for DynamicViT-S and our DeiT-S+PSS model. Note that $\rho$ does not have exactly the same meaning for each model, as DynamicVit has multiple sparsification stages. PSS achieves a more gradual trade-off between accuracy and $\rho$.}
    \label{fig:dynamicvit-comp}
\end{figure}

We also tried to see how well DynamicVit~\cite{rao2021dynamicvit} would perform under different patch keep rates $\rho$. Note that, unlike our method, DynamicViT aims to hit a target patch sparsification setting and is not really intended to support a dynamic inference range. Additionally, the $\rho$ parameter works a bit differently for DynamicVit as it uses multiple stages of sparsification. Figure~\ref{fig:dynamicvit-comp} shows the dynamic inference performance of DieT-S (224$\times$224 image size) trained under DynamicViT and a cyclic PSS. One interesting observation is that DynamicViT has essentially zero improvement in validation accuracy in the $\rho=1.0$ to $\rho=0.8$ range compared to the target of $\rho=0.7$. We believe this makes sense, as DynamicVit was trained to with additional loss term to support the specific target of $\rho=0.7$. By comparison, our cyclic PSS model is trained with a variety of $\rho$ settings and therefore achieves a more gradual trade-off between accuracy and the patch keep rate (corresponding to image throughput).

\subsection{Comparing Patch Selection of ViT Models}
\begin{figure*}
    \includegraphics[width=\textwidth, height=0.48\textheight]{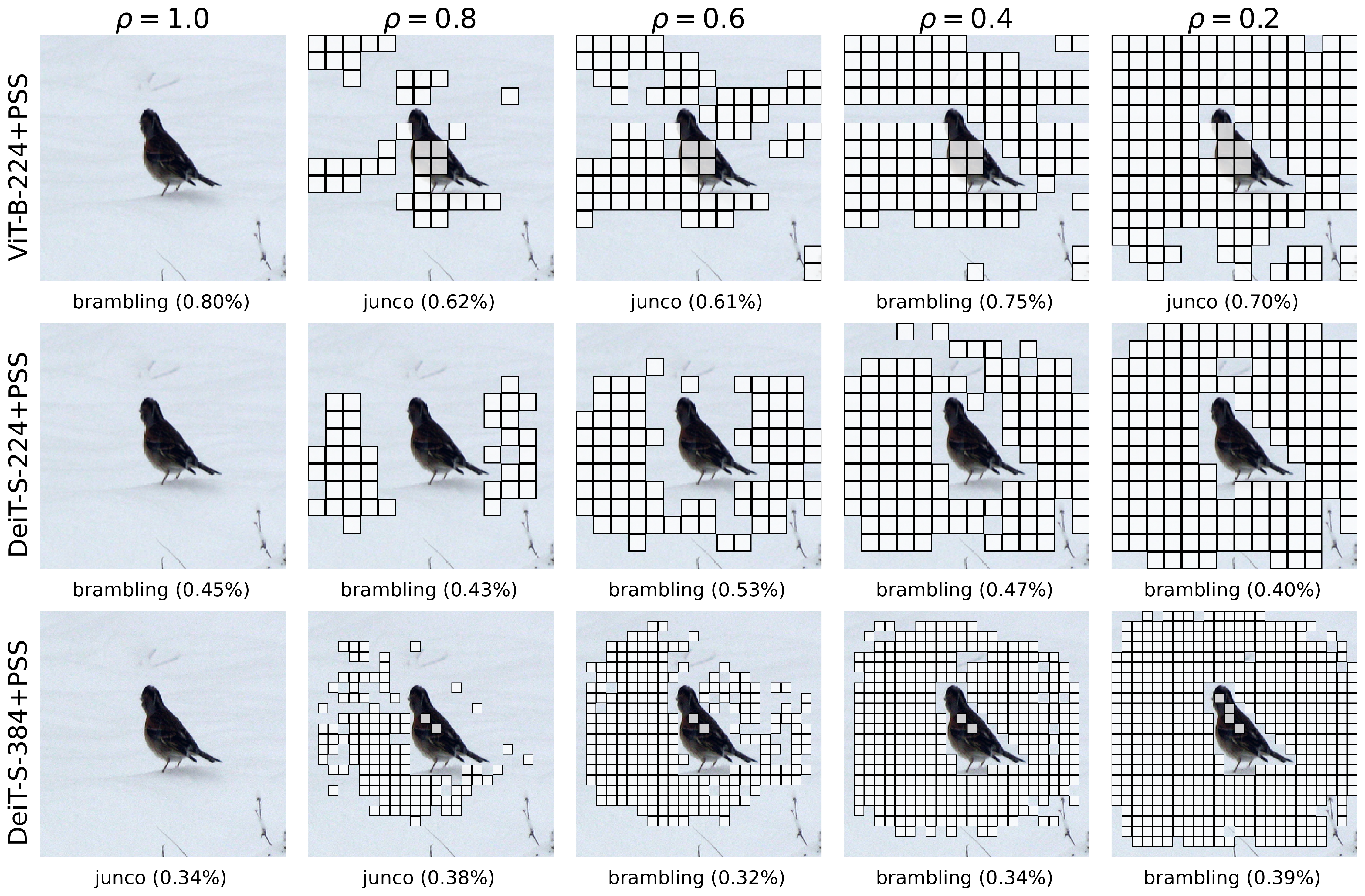}        \includegraphics[width=\textwidth, height=0.48\textheight]{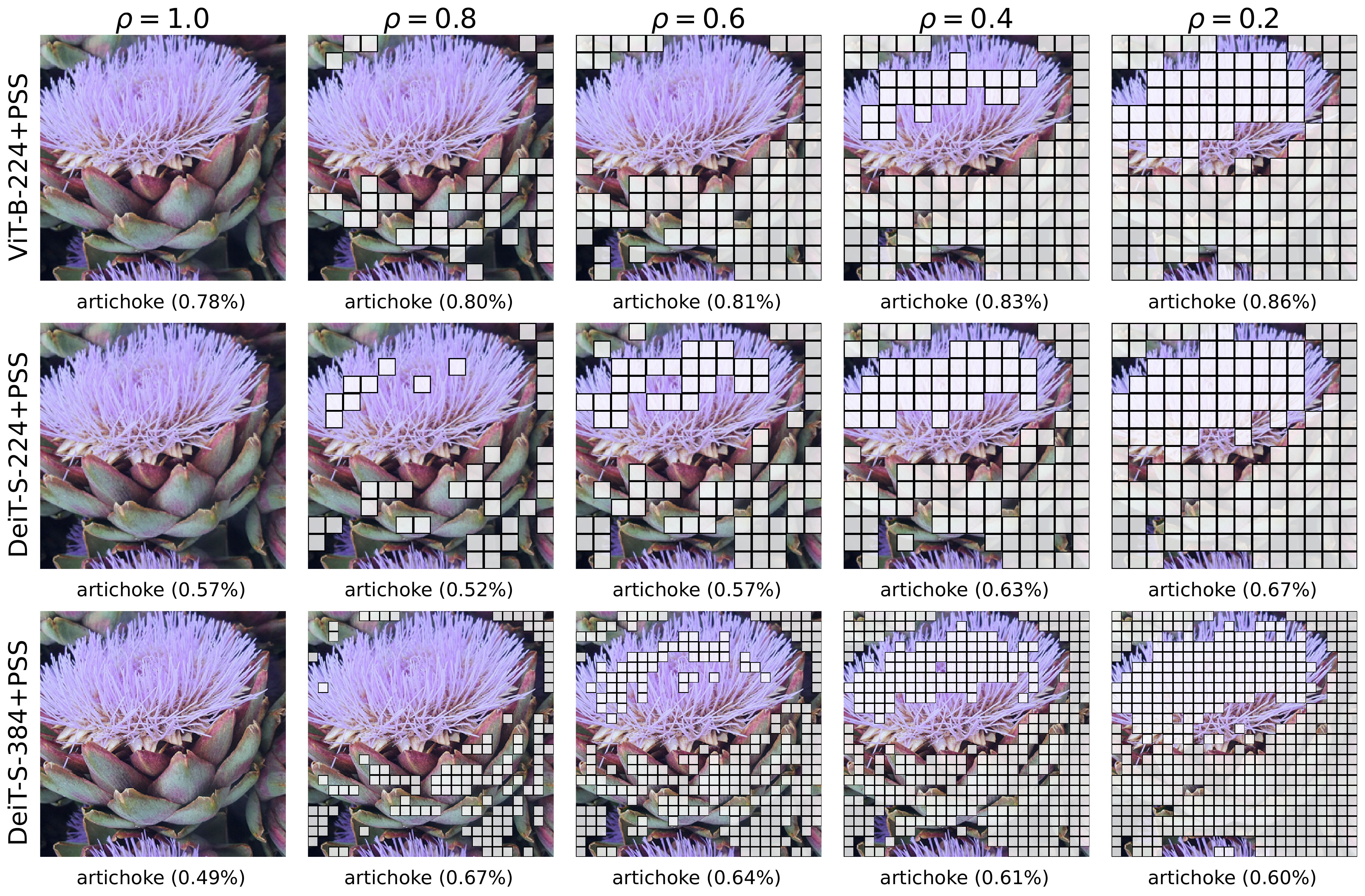}
    \caption{(top) Brambling. (bottom) Artichoke.}
    \label{fig:image-patches-comp-a}
\end{figure*}

\begin{figure*}
    \includegraphics[width=\textwidth, height=0.48\textheight]{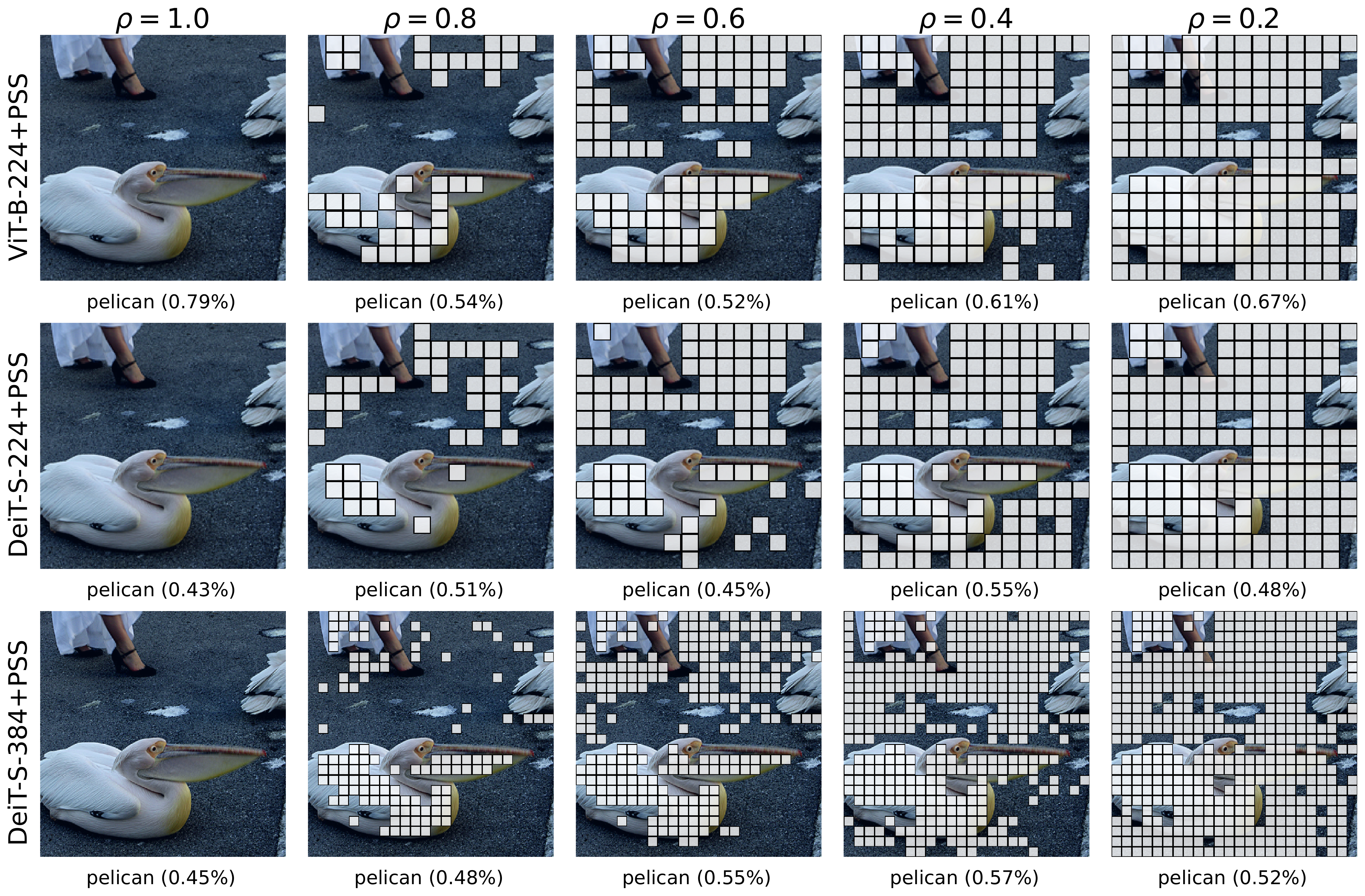}        \includegraphics[width=\textwidth, height=0.48\textheight]{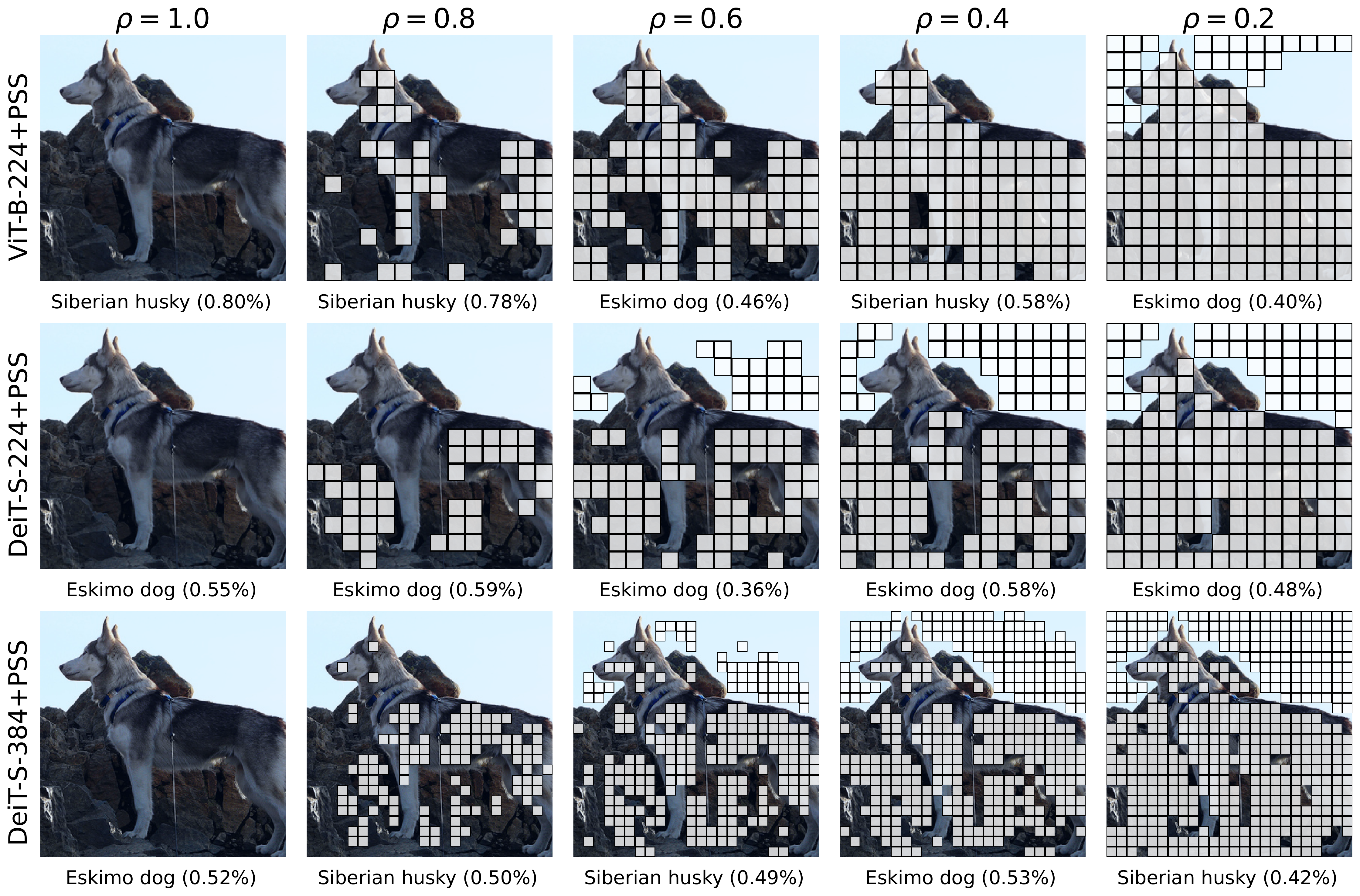}
    \caption{(top) Pelican. (bottom) Siberian Husky.}
    \label{fig:image-patches-comp-b}
\end{figure*}

In this section, we compare the patch selection for three PSS models: ViT-B-224, DieT-S-224, and DieT-S-384. All models are trained using a cyclic PSS and a magnitude PSF. Figures~\ref{fig:image-patches-comp-a}~and~\ref{fig:image-patches-comp-b} show which patches each of the three models select for varying values of $\rho$. The text below each image is the predicted class with probability in parenthesis for that combination of a given model and $\rho$ setting. For the DieT-S-384 rows, the dropped patches appear smaller due to the images being downsampled (from 384 to 224) to better fit the figure. For both images sizes, we use a patch size of 16 $\times$ 16, leading to 24$\times$24 patches for the 384$\times$384 resolution and 14$\times$14 patches for the 224$\times$224 resolution.

Looking at the figures, we note some fluctuation across the selected class, especially as $\rho$ is decreased. For instance, Figure~\ref{fig:image-patches-comp-a} (top) is an image of a brambling, however the ViT-B model changes its prediction to a junco (another type of bird) for many of the smaller $\rho$ settings. Generally, as fewer patches are selected, we observe that all models tend to keep patches on the boundary of the object of classification and the background. Often, the middle portion of the classification target has patches that have been dropped.

\end{document}